\documentclass[runningheads]{llncs}
\usepackage[T1]{fontenc}
\usepackage{graphicx}
\usepackage{cite}
\usepackage{amsmath,amssymb,amsfonts}
\usepackage{algorithmic}
\usepackage{graphicx}
\usepackage{textcomp}
\usepackage{multirow}
\usepackage{booktabs}
\usepackage{xcolor}
\usepackage{tikz}
\begin{document}
\title{A Novel Bounding Box Regression Method for Single Object Tracking}
%
%
\author{Omar Abdelaziz\inst{1}\orcidID{0009-0009-2416-5700} \and
Mohmaed Sami Shehata\inst{1}\orcidID{0000-0003-1289-0285}}
%
%
\institute{University of British Columbia, Kelowna BC V1V1V7, Canada 
\email{oabdelaz@student.ubc.ca}\\}
\maketitle              
\begin{abstract}
Locating an object in a sequence of frames, given its appearance in the first frame of the sequence, is a hard problem that involves many stages. Usually, state-of-the-art methods focus on bringing novel ideas in the visual encoding or relational modelling phases. However, in this work, we show that bounding box regression from learned joint search and template features is of high importance as well. While previous methods relied heavily on well-learned features representing interactions between search and template, we hypothesize that the receptive field of the input convolutional bounding box network plays an important role in accurately determining the object location. To this end, we introduce two novel bounding box regression networks: inception and deformable. Experiments and ablation studies show that our inception module installed on the recent ODTrack outperforms the latter on three benchmarks: the GOT-10k, the UAV123 and the OTB2015. 

\end{abstract}

\section{Introduction}
Single object tracking is a challenging computer vision subfield that is ubiquitous in many applications such as video surveillance and augmented reality. The challenging nature of tracking is associated with the problem definition as well as the techniques to address the problem. Given a sequence of images, it is required to identify the size and location of an object given by a bounding box annotation in the first frame of the sequence. In single-object tracking, the object is also identified by a bounding box. 

The current prevalent pipeline in single object tracking is the joint feature extraction and relation modelling pipeline introduced in \cite{cui2022mixformer} and \cite{ye2022joint}. Exploiting Vision Transformers (ViTs), this pipeline leverages the natural attention capabilities in relationship association to locate the template object in the search frame in a low-dimensional space. A bounding box prediction head is used afterwards to map the transformer attention map into bounding box coordinates. 

Despite showing remarkable success, the current tracking methods are focused on the feature extraction and relation modelling phases in a one-stream manner while being less attentive to the bounding box regression phase. Bounding box regression is usually done using a simple multi-branch convolutional network. These networks simplify the bounding box regression by dividing it into classification and regression subproblems. The classification stage aims to separate the object from the background. In contrast, the regression branch aims to predict four float values that represent the coordinates of the top-left corner and either the width and the height or the coordinates of the bottom-left corner of the box. This simple design depends on the number of parameters of the Vision Transformer \cite{dosovitskiy2021an} to learn a good embedded representation that overcomes problems in the bounding box network. 

We hypothesize that the simple convolutional pipeline of bounding box networks is not enough to capture complex global and local relations in the embedded feature representation output by the transformer. We introduce two simple yet effective bounding box networks that are capable of learning a variable receptive field to enable a more accurate determination of the target location. In particular, we introduce the Inception and the deformable convolution Inception bounding box networks based on the Inception \cite{szegedy2015going} and deformable convolutional concepts \cite{Zhu2019Deformable}. We show backed by experiments that our bounding box networks can capture more useful information from the ViT embeddings. 

To this end, the contributions of this work are three-fold:
\begin{itemize}
    \item We propose a new bounding box head for ViT-based single object trackers, namely, the Inception module that addresses the fixed local receptive field of the bounding box network of single object tracking transformers. 
    \item We propose another variant of the Inception module, the deformable convolution Inception module, which learns the required receptive field for a better-performing bounding box network.
    \item We conduct extensive experiments and ablation studies to show the effectiveness of our proposed modules. Our best model sets a new state-of-the-art. 
\end{itemize}

\section{Related Works}
\subsection{Tracking Paradigms} Recent deep-learning-based single-object trackers involve either CNN or ViT backbones. CNN paradigm has been dominated by the Siamese architecture since their introduction in \cite{bertinetto2016fully, Tao2016Siamese}. The basic idea of cross-correlation in the embedding space employed in \cite{bertinetto2016fully} is employed in many variants \cite{Chen2020Siamese, Li2018High, Li2019SiamRPN++, Wang2019Fast}. Siamese trackers aim to embed the template object and the search image in a well-chosen space where the cross-correlation between the embedded representations results in a score map representing the object's location. Finally, post-processing techniques such as the cosine window penalty have to be adopted to calculate the bounding box coordinates from the correlation score map. Despite showing remarkable success, this paradigm had several drawbacks. First, the simple cross-correlation operation assumes high-quality feature embeddings, and this is not guaranteed by simple CNN as AlexNet \cite{Krizhevsky2012ImageNet} or VGGNet \cite{Simonyan2015Very} that were used as the feature extractor. Second, the complex post-processing steps required in the bounding box coordinates calculation phase may hinder the performance if its parameters are not chosen optimally. In our work, we hypothesize that bounding box coordinates must be learned as a part of an end-to-end learning framework without a hand-crafted set of thresholds or parameters to achieve the robustness and accuracy of bounding box calculation. Our method not only learns the coordinates of the bounding box directly but also the receptive field of the bounding box CNN, thus, resulting in extracting more useful information from the previous phase.

The CNN paradigm was dominant until the ViT-based trackers were introduced \cite{chen2021transformer, Wang2021Transformer}. These trackers reimplemented the dominant Siamese tracking with a transformer visual encoder.  Other research leveraged the powerful attention capabilities to bring closer end-to-end ViT tracking. Several works simultaneously introduced joint feature extraction and relation modelling \cite{Boyu2022Backbone, ye2022joint, cui2022mixformer}. Generally, this framework assumes that the transformer is capable of extracting the visual features alongside learning the relationships between the embedded representations. After obtaining a representative feature map from the ViT, these methods apply a convolutional bounding box head to predict the bounding box coordinates mainly in the classification and regression phases. This one-stream paradigm has been dominant with many works that propose changes to ViT architecture itself to adapt to the single object tracking problem \cite{Shenyuan2022AiATrack, Song2022Transformer, Chen2023SeqTrack, wei2023autoregressive}.

\subsection{Convolutional Design Choices}
\subsubsection{Inception network} The inception network was introduced in \cite{Simonyan2015Very} to overcome the problem of the narrow receptive field of previous CNNs. The main idea behind the original paper and its variants \cite{Szegedy2016Rethinking} is that for a convolutional network to go deeper, its receptive field has to be wide enough to process the input on many deep stages without losing information from the early stages. For that purpose, the Inception module carries on several convolutional passes in parallel, each representing a different convolutional filter. Different outputs of the passes are concatenated together and passed to a $1 \times 1$ convolutional filter to reduce the channel dimension in a learnable manner. In this work, we apply the simplest inception module to the output of a ViT encoder from three known trackers. 

\subsubsection{Deformable Convolutions} Deformable convolution \cite{Dai2017Deformable} is an extension of the regular convolution where the sampling positions of the convolutional filter are learnable. In this regard, the deformable convolution operation involves two runs of convolutional filters. The first run determines the offsets that the filter of the second run will use to sample values of the input feature map, which will apply the convolution operation. In another version of the deformable convolutions \cite{Zhu2019Deformable}, the amplitudes of the input feature maps are modulated by a learnable mask, ensuring an extra degree of freedom. We implement a variant of the proposed Inception bounding box network with deformable convolutions to exploit the powerful learning paradigm encapsulated by the learnable sampling positions and input amplitudes. 

\section{Methodology}
\label{sec:methodology}
  
\tikzset {_z6nr22wdl/.code = {\pgfsetadditionalshadetransform{ \pgftransformshift{\pgfpoint{89.1 bp } { -108.9 bp }  }  \pgftransformscale{1.32 }  }}}
\pgfdeclareradialshading{_rvrfkfnsm}{\pgfpoint{-72bp}{88bp}}{rgb(0bp)=(1,1,1);
rgb(0bp)=(1,1,1);
rgb(25bp)=(0,0,0);
rgb(400bp)=(0,0,0)}

  
\tikzset {_ksfhplt3j/.code = {\pgfsetadditionalshadetransform{ \pgftransformshift{\pgfpoint{89.1 bp } { -108.9 bp }  }  \pgftransformscale{1.32 }  }}}
\pgfdeclareradialshading{_308liggkf}{\pgfpoint{-72bp}{88bp}}{rgb(0bp)=(1,1,1);
rgb(0bp)=(1,1,1);
rgb(25bp)=(0,0,0);
rgb(400bp)=(0,0,0)}

  
\tikzset {_e7chbpljq/.code = {\pgfsetadditionalshadetransform{ \pgftransformshift{\pgfpoint{89.1 bp } { -108.9 bp }  }  \pgftransformscale{1.32 }  }}}
\pgfdeclareradialshading{_htnhudn2p}{\pgfpoint{-72bp}{88bp}}{rgb(0bp)=(1,1,1);
rgb(0bp)=(1,1,1);
rgb(25bp)=(0,0,0);
rgb(400bp)=(0,0,0)}

\tikzset{every picture/.style={line width=0.75pt}} 

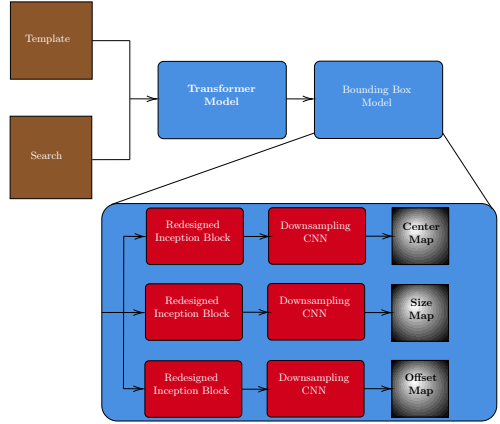
\begin{figure}[ht]
    \centering
    \scalebox{0.45}{ 
        \begin{tikzpicture}[x=0.75pt,y=0.75pt,yscale=-1,xscale=1]
        
        \draw  [fill={rgb, 255:red, 139; green, 87; blue, 42 }  ,fill opacity=1 ] (77.72,43.13) -- (169.35,43.13) -- (169.35,129.85) -- (77.72,129.85) -- cycle ;
        \draw  [fill={rgb, 255:red, 139; green, 87; blue, 42 }  ,fill opacity=1 ] (77.39,171.35) -- (168.89,171.35) -- (168.89,263.85) -- (77.39,263.85) -- cycle ;
        \draw  [fill={rgb, 255:red, 74; green, 144; blue, 226 }  ,fill opacity=1 ] (243.22,116.06) .. controls (243.22,112.63) and (246,109.85) .. (249.43,109.85) -- (380.68,109.85) .. controls (384.11,109.85) and (386.89,112.63) .. (386.89,116.06) -- (386.89,190.39) .. controls (386.89,193.82) and (384.11,196.6) .. (380.68,196.6) -- (249.43,196.6) .. controls (246,196.6) and (243.22,193.82) .. (243.22,190.39) -- cycle ;
        \draw    (211.84,151.85) -- (239.49,151.94) ;
        \draw [shift={(241.49,151.95)}, rotate = 180.19] [color={rgb, 255:red, 0; green, 0; blue, 0 }  ][line width=0.75]    (10.93,-3.29) .. controls (6.95,-1.4) and (3.31,-0.3) .. (0,0) .. controls (3.31,0.3) and (6.95,1.4) .. (10.93,3.29)   ;
        \draw    (169.34,220.1) -- (211.54,220.05) ;
        \draw [fill={rgb, 255:red, 139; green, 87; blue, 42 }  ,fill opacity=1 ]   (169,86.63) -- (211.54,86.6) ;
        \draw    (211.54,86.6) -- (211.54,220.05) ;
        \draw  [fill={rgb, 255:red, 74; green, 144; blue, 226 }  ,fill opacity=1 ] (418.47,115.81) .. controls (418.47,112.38) and (421.25,109.6) .. (424.68,109.6) -- (555.93,109.6) .. controls (559.36,109.6) and (562.14,112.38) .. (562.14,115.81) -- (562.14,190.14) .. controls (562.14,193.57) and (559.36,196.35) .. (555.93,196.35) -- (424.68,196.35) .. controls (421.25,196.35) and (418.47,193.57) .. (418.47,190.14) -- cycle ;
        
        \draw    (386.66,151.9) -- (414.31,151.99) ;
        \draw [shift={(416.31,152)}, rotate = 180.19] [color={rgb, 255:red, 0; green, 0; blue, 0 }  ][line width=0.75]    (10.93,-3.29) .. controls (6.95,-1.4) and (3.31,-0.3) .. (0,0) .. controls (3.31,0.3) and (6.95,1.4) .. (10.93,3.29)   ;
        \draw  [fill={rgb, 255:red, 74; green, 144; blue, 226 }  ,fill opacity=1 ] (180.13,286.79) .. controls (180.13,277.15) and (187.94,269.34) .. (197.58,269.34) -- (605.35,269.34) .. controls (614.99,269.34) and (622.8,277.15) .. (622.8,286.79) -- (622.8,495.79) .. controls (622.8,505.42) and (614.99,513.24) .. (605.35,513.24) -- (197.58,513.24) .. controls (187.94,513.24) and (180.13,505.42) .. (180.13,495.79) -- cycle ;
        \draw    (180.33,391.5) -- (204.58,391.5) ;
        \draw    (204.77,306.02) -- (204.4,476.98) ;
        \draw    (204.77,306.02) -- (225.94,306.32) ;
        \draw [shift={(227.94,306.35)}, rotate = 180.82] [color={rgb, 255:red, 0; green, 0; blue, 0 }  ][line width=0.75]    (10.93,-3.29) .. controls (6.95,-1.4) and (3.31,-0.3) .. (0,0) .. controls (3.31,0.3) and (6.95,1.4) .. (10.93,3.29)   ;
        \draw    (204.58,391.5) -- (224.5,391.65) ;
        \draw [shift={(226.5,391.67)}, rotate = 180.44] [color={rgb, 255:red, 0; green, 0; blue, 0 }  ][line width=0.75]    (10.93,-3.29) .. controls (6.95,-1.4) and (3.31,-0.3) .. (0,0) .. controls (3.31,0.3) and (6.95,1.4) .. (10.93,3.29)   ;
        \draw    (204.4,476.98) -- (224.31,477.13) ;
        \draw [shift={(226.31,477.15)}, rotate = 180.44] [color={rgb, 255:red, 0; green, 0; blue, 0 }  ][line width=0.75]    (10.93,-3.29) .. controls (6.95,-1.4) and (3.31,-0.3) .. (0,0) .. controls (3.31,0.3) and (6.95,1.4) .. (10.93,3.29)   ;
        
        \path  [shading=_rvrfkfnsm,_z6nr22wdl] (505.46,274.25) -- (568.96,274.25) -- (568.96,337.75) -- (505.46,337.75) -- cycle ; 
         \draw   (505.46,274.25) -- (568.96,274.25) -- (568.96,337.75) -- (505.46,337.75) -- cycle ; 
        
        \draw  [fill={rgb, 255:red, 208; green, 2; blue, 27 }  ,fill opacity=1 ] (230.08,278.89) .. controls (230.08,276.38) and (232.11,274.35) .. (234.62,274.35) -- (334.18,274.35) .. controls (336.69,274.35) and (338.72,276.38) .. (338.72,278.89) -- (338.72,333.21) .. controls (338.72,335.72) and (336.69,337.75) .. (334.18,337.75) -- (234.62,337.75) .. controls (232.11,337.75) and (230.08,335.72) .. (230.08,333.21) -- cycle ;
        
        \draw  [fill={rgb, 255:red, 208; green, 2; blue, 27 }  ,fill opacity=1 ] (367.08,278.55) .. controls (367.08,276.05) and (369.11,274.02) .. (371.62,274.02) -- (471.18,274.02) .. controls (473.69,274.02) and (475.72,276.05) .. (475.72,278.55) -- (475.72,332.88) .. controls (475.72,335.39) and (473.69,337.42) .. (471.18,337.42) -- (371.62,337.42) .. controls (369.11,337.42) and (367.08,335.39) .. (367.08,332.88) -- cycle ;
        
        \path  [shading=_308liggkf,_ksfhplt3j] (505.21,360.25) -- (568.71,360.25) -- (568.71,423.75) -- (505.21,423.75) -- cycle ; 
         \draw   (505.21,360.25) -- (568.71,360.25) -- (568.71,423.75) -- (505.21,423.75) -- cycle ; 
        
        \path  [shading=_htnhudn2p,_e7chbpljq] (505.21,445.25) -- (568.71,445.25) -- (568.71,508.75) -- (505.21,508.75) -- cycle ; 
         \draw   (505.21,445.25) -- (568.71,445.25) -- (568.71,508.75) -- (505.21,508.75) -- cycle ; 
        
        \draw    (476,305.9) -- (503.65,305.99) ;
        \draw [shift={(505.65,306)}, rotate = 180.19] [color={rgb, 255:red, 0; green, 0; blue, 0 }  ][line width=0.75]    (10.93,-3.29) .. controls (6.95,-1.4) and (3.31,-0.3) .. (0,0) .. controls (3.31,0.3) and (6.95,1.4) .. (10.93,3.29)   ;
        \draw    (474.85,391.57) -- (502.5,391.66) ;
        \draw [shift={(504.5,391.67)}, rotate = 180.19] [color={rgb, 255:red, 0; green, 0; blue, 0 }  ][line width=0.75]    (10.93,-3.29) .. controls (6.95,-1.4) and (3.31,-0.3) .. (0,0) .. controls (3.31,0.3) and (6.95,1.4) .. (10.93,3.29)   ;
        \draw    (474.4,476.98) -- (502.46,477.08) ;
        \draw [shift={(504.46,477.08)}, rotate = 180.19] [color={rgb, 255:red, 0; green, 0; blue, 0 }  ][line width=0.75]    (10.93,-3.29) .. controls (6.95,-1.4) and (3.31,-0.3) .. (0,0) .. controls (3.31,0.3) and (6.95,1.4) .. (10.93,3.29)   ;
        \draw    (418.47,190.14) -- (190.63,270.5) ;
        \draw    (562.14,190.14) -- (617.8,275.3) ;
        \draw    (338.35,306.19) -- (366,306.28) ;
        \draw [shift={(368,306.29)}, rotate = 180.19] [color={rgb, 255:red, 0; green, 0; blue, 0 }  ][line width=0.75]    (10.93,-3.29) .. controls (6.95,-1.4) and (3.31,-0.3) .. (0,0) .. controls (3.31,0.3) and (6.95,1.4) .. (10.93,3.29)   ;
        \draw  [fill={rgb, 255:red, 208; green, 2; blue, 27 }  ,fill opacity=1 ] (229.08,363.89) .. controls (229.08,361.38) and (231.11,359.35) .. (233.62,359.35) -- (333.18,359.35) .. controls (335.69,359.35) and (337.72,361.38) .. (337.72,363.89) -- (337.72,418.21) .. controls (337.72,420.72) and (335.69,422.75) .. (333.18,422.75) -- (233.62,422.75) .. controls (231.11,422.75) and (229.08,420.72) .. (229.08,418.21) -- cycle ;
        
        \draw  [fill={rgb, 255:red, 208; green, 2; blue, 27 }  ,fill opacity=1 ] (366.08,363.55) .. controls (366.08,361.05) and (368.11,359.02) .. (370.62,359.02) -- (470.18,359.02) .. controls (472.69,359.02) and (474.72,361.05) .. (474.72,363.55) -- (474.72,417.88) .. controls (474.72,420.39) and (472.69,422.42) .. (470.18,422.42) -- (370.62,422.42) .. controls (368.11,422.42) and (366.08,420.39) .. (366.08,417.88) -- cycle ;
        
        \draw    (337.35,391.19) -- (365,391.28) ;
        \draw [shift={(367,391.29)}, rotate = 180.19] [color={rgb, 255:red, 0; green, 0; blue, 0 }  ][line width=0.75]    (10.93,-3.29) .. controls (6.95,-1.4) and (3.31,-0.3) .. (0,0) .. controls (3.31,0.3) and (6.95,1.4) .. (10.93,3.29)   ;
        \draw  [fill={rgb, 255:red, 208; green, 2; blue, 27 }  ,fill opacity=1 ] (228.42,450.22) .. controls (228.42,447.71) and (230.45,445.68) .. (232.95,445.68) -- (332.52,445.68) .. controls (335.02,445.68) and (337.05,447.71) .. (337.05,450.22) -- (337.05,504.55) .. controls (337.05,507.05) and (335.02,509.08) .. (332.52,509.08) -- (232.95,509.08) .. controls (230.45,509.08) and (228.42,507.05) .. (228.42,504.55) -- cycle ;
        
        \draw  [fill={rgb, 255:red, 208; green, 2; blue, 27 }  ,fill opacity=1 ] (365.42,449.89) .. controls (365.42,447.38) and (367.45,445.35) .. (369.95,445.35) -- (469.52,445.35) .. controls (472.02,445.35) and (474.05,447.38) .. (474.05,449.89) -- (474.05,504.21) .. controls (474.05,506.72) and (472.02,508.75) .. (469.52,508.75) -- (369.95,508.75) .. controls (367.45,508.75) and (365.42,506.72) .. (365.42,504.21) -- cycle ;
        
        \draw    (336.69,477.52) -- (364.34,477.61) ;
        \draw [shift={(366.34,477.62)}, rotate = 180.19] [color={rgb, 255:red, 0; green, 0; blue, 0 }  ][line width=0.75]    (10.93,-3.29) .. controls (6.95,-1.4) and (3.31,-0.3) .. (0,0) .. controls (3.31,0.3) and (6.95,1.4) .. (10.93,3.29)   ;
        
        \draw (98.92,209.1) node [anchor=north west][inner sep=0.75pt]   [align=left] {\textcolor[rgb]{1,1,1}{Search}};
        \draw (92.83,78.69) node [anchor=north west][inner sep=0.75pt]   [align=left] {\textcolor[rgb]{1,1,1}{Template}};
        \draw (274.23,134.86) node [anchor=north west][inner sep=0.75pt]  [color={rgb, 255:red, 255; green, 255; blue, 255 }  ,opacity=1 ] [align=left] {\begin{minipage}[lt]{57.89pt}\setlength\topsep{0pt}
        \begin{center}
        \textbf{Transformer}\\\textcolor[rgb]{1,1,1}{\textbf{Model}}
        \end{center}
        
        \end{minipage}};
        \draw (442.73,135.11) node [anchor=north west][inner sep=0.75pt]  [color={rgb, 255:red, 255; green, 255; blue, 255 }  ,opacity=1 ] [align=left] {\begin{minipage}[lt]{66.24pt}\setlength\topsep{0pt}
        \begin{center}
        Bounding Box\\\textcolor[rgb]{1,1,1}{Model}
        \end{center}
        
        \end{minipage}};
        \draw (371.85,287.55) node [anchor=north west][inner sep=0.75pt]  [color={rgb, 255:red, 255; green, 255; blue, 255 }  ,opacity=1 ] [align=left] {\begin{minipage}[lt]{69.63pt}\setlength\topsep{0pt}
        \begin{center}
        Downsampling\\CNN
        \end{center}
        
        \end{minipage}};
        \draw (232.85,286.88) node [anchor=north west][inner sep=0.75pt]  [color={rgb, 255:red, 255; green, 255; blue, 255 }  ,opacity=1 ] [align=left] {\begin{minipage}[lt]{71.9pt}\setlength\topsep{0pt}
        \begin{center}
        Redesigned\\Inception Block
        \end{center}
        
        \end{minipage}};
        \draw (370.85,372.55) node [anchor=north west][inner sep=0.75pt]  [color={rgb, 255:red, 255; green, 255; blue, 255 }  ,opacity=1 ] [align=left] {\begin{minipage}[lt]{69.63pt}\setlength\topsep{0pt}
        \begin{center}
        Downsampling\\CNN
        \end{center}
        
        \end{minipage}};
        \draw (231.85,371.88) node [anchor=north west][inner sep=0.75pt]  [color={rgb, 255:red, 255; green, 255; blue, 255 }  ,opacity=1 ] [align=left] {\begin{minipage}[lt]{71.9pt}\setlength\topsep{0pt}
        \begin{center}
        Redesigned\\Inception Block
        \end{center}
        
        \end{minipage}};
        \draw (370.18,458.88) node [anchor=north west][inner sep=0.75pt]  [color={rgb, 255:red, 255; green, 255; blue, 255 }  ,opacity=1 ] [align=left] {\begin{minipage}[lt]{69.63pt}\setlength\topsep{0pt}
        \begin{center}
        Downsampling\\CNN
        \end{center}
        
        \end{minipage}};
        \draw (231.18,458.21) node [anchor=north west][inner sep=0.75pt]  [color={rgb, 255:red, 255; green, 255; blue, 255 }  ,opacity=1 ] [align=left] {\begin{minipage}[lt]{71.9pt}\setlength\topsep{0pt}
        \begin{center}
        Redesigned\\Inception Block
        \end{center}
        
        \end{minipage}};
        \draw (513.5,288.7) node [anchor=north west][inner sep=0.75pt]  [color={rgb, 255:red, 0; green, 0; blue, 0 }  ,opacity=1 ] [align=left] {\begin{minipage}[lt]{35.03pt}\setlength\topsep{0pt}
        \begin{center}
        \textbf{Center}\\\textbf{Map}
        \end{center}
        
        \end{minipage}};
        \draw (521.5,374.7) node [anchor=north west][inner sep=0.75pt]  [color={rgb, 255:red, 0; green, 0; blue, 0 }  ,opacity=1 ] [align=left] {\begin{minipage}[lt]{23.13pt}\setlength\topsep{0pt}
        \begin{center}
        \textbf{Size}\\\textbf{Map}
        \end{center}
        
        \end{minipage}};
        \draw (515,458.2) node [anchor=north west][inner sep=0.75pt]  [color={rgb, 255:red, 0; green, 0; blue, 0 }  ,opacity=1 ] [align=left] {\begin{minipage}[lt]{32.19pt}\setlength\topsep{0pt}
        \begin{center}
        \textbf{Offset}\\\textbf{Map}
        \end{center}
        
        \end{minipage}};

        \end{tikzpicture}
    }
    \caption{The overall architecture of the recent ViT-based single object trackers.}
    \label{fig:overall_arch}
\end{figure}

\begin{figure}[ht]
    \centering 
    \scalebox{0.4}{

\tikzset{every picture/.style={line width=0.75pt}} 

\begin{tikzpicture}[x=0.75pt,y=0.75pt,yscale=-1,xscale=1]

\draw  [fill={rgb, 255:red, 208; green, 2; blue, 27 }  ,fill opacity=1 ] (56.91,37.62) .. controls (56.91,22.02) and (69.56,9.36) .. (85.17,9.36) -- (553.84,9.36) .. controls (569.45,9.36) and (582.1,22.02) .. (582.1,37.62) -- (582.1,376.06) .. controls (582.1,391.67) and (569.45,404.32) .. (553.84,404.32) -- (85.17,404.32) .. controls (69.56,404.32) and (56.91,391.67) .. (56.91,376.06) -- cycle ;
\draw  [fill={rgb, 255:red, 0; green, 91; blue, 198 }  ,fill opacity=1 ] (104.97,55.53) -- (184.53,55.53) -- (184.53,88.87) -- (104.97,88.87) -- cycle ;
\draw  [fill={rgb, 255:red, 0; green, 91; blue, 198 }  ,fill opacity=1 ] (416.7,208.18) .. controls (416.7,197.2) and (425.6,188.3) .. (436.58,188.3) .. controls (447.55,188.3) and (456.45,197.2) .. (456.45,208.18) .. controls (456.45,219.15) and (447.55,228.05) .. (436.58,228.05) .. controls (425.6,228.05) and (416.7,219.15) .. (416.7,208.18) -- cycle ;
\draw [fill={rgb, 255:red, 0; green, 91; blue, 198 }  ,fill opacity=1 ]   (289,344.07) -- (435,229.29) ;
\draw [shift={(436.58,228.05)}, rotate = 141.83] [color={rgb, 255:red, 0; green, 0; blue, 0 }  ][line width=0.75]    (10.93,-3.29) .. controls (6.95,-1.4) and (3.31,-0.3) .. (0,0) .. controls (3.31,0.3) and (6.95,1.4) .. (10.93,3.29)   ;
\draw [fill={rgb, 255:red, 0; green, 91; blue, 198 }  ,fill opacity=1 ]   (288.33,252.07) -- (422.67,224.45) ;
\draw [shift={(424.63,224.05)}, rotate = 168.38] [color={rgb, 255:red, 0; green, 0; blue, 0 }  ][line width=0.75]    (10.93,-3.29) .. controls (6.95,-1.4) and (3.31,-0.3) .. (0,0) .. controls (3.31,0.3) and (6.95,1.4) .. (10.93,3.29)   ;
\draw [fill={rgb, 255:red, 0; green, 91; blue, 198 }  ,fill opacity=1 ]   (185.67,162.07) -- (420.89,192.54) ;
\draw [shift={(422.88,192.8)}, rotate = 187.38] [color={rgb, 255:red, 0; green, 0; blue, 0 }  ][line width=0.75]    (10.93,-3.29) .. controls (6.95,-1.4) and (3.31,-0.3) .. (0,0) .. controls (3.31,0.3) and (6.95,1.4) .. (10.93,3.29)   ;
\draw [fill={rgb, 255:red, 0; green, 91; blue, 198 }  ,fill opacity=1 ]   (391.67,70.73) -- (435.86,186.43) ;
\draw [shift={(436.58,188.3)}, rotate = 249.09] [color={rgb, 255:red, 0; green, 0; blue, 0 }  ][line width=0.75]    (10.93,-3.29) .. controls (6.95,-1.4) and (3.31,-0.3) .. (0,0) .. controls (3.31,0.3) and (6.95,1.4) .. (10.93,3.29)   ;
\draw    (57.33,207.87) -- (81.58,207.87) ;
\draw    (81.77,71.5) -- (81.4,344.23) ;
\draw    (81.77,71.5) -- (102,71.7) ;
\draw [shift={(104,71.71)}, rotate = 180.55] [color={rgb, 255:red, 0; green, 0; blue, 0 }  ][line width=0.75]    (10.93,-3.29) .. controls (6.95,-1.4) and (3.31,-0.3) .. (0,0) .. controls (3.31,0.3) and (6.95,1.4) .. (10.93,3.29)   ;
\draw    (81.58,162.87) -- (101.5,163.11) ;
\draw [shift={(103.5,163.13)}, rotate = 180.7] [color={rgb, 255:red, 0; green, 0; blue, 0 }  ][line width=0.75]    (10.93,-3.29) .. controls (6.95,-1.4) and (3.31,-0.3) .. (0,0) .. controls (3.31,0.3) and (6.95,1.4) .. (10.93,3.29)   ;
\draw    (81.4,344.23) -- (101.43,344.28) ;
\draw [shift={(103.43,344.29)}, rotate = 180.13] [color={rgb, 255:red, 0; green, 0; blue, 0 }  ][line width=0.75]    (10.93,-3.29) .. controls (6.95,-1.4) and (3.31,-0.3) .. (0,0) .. controls (3.31,0.3) and (6.95,1.4) .. (10.93,3.29)   ;
\draw    (81.87,253.15) -- (101.79,253.39) ;
\draw [shift={(103.79,253.42)}, rotate = 180.7] [color={rgb, 255:red, 0; green, 0; blue, 0 }  ][line width=0.75]    (10.93,-3.29) .. controls (6.95,-1.4) and (3.31,-0.3) .. (0,0) .. controls (3.31,0.3) and (6.95,1.4) .. (10.93,3.29)   ;
\draw    (560.18,208.01) -- (580.1,208.25) ;
\draw [shift={(582.1,208.28)}, rotate = 180.7] [color={rgb, 255:red, 0; green, 0; blue, 0 }  ][line width=0.75]    (10.93,-3.29) .. controls (6.95,-1.4) and (3.31,-0.3) .. (0,0) .. controls (3.31,0.3) and (6.95,1.4) .. (10.93,3.29)   ;
\draw  [fill={rgb, 255:red, 0; green, 91; blue, 198 }  ,fill opacity=1 ] (104.63,146.53) -- (184.2,146.53) -- (184.2,179.87) -- (104.63,179.87) -- cycle ;

\draw  [fill={rgb, 255:red, 0; green, 91; blue, 198 }  ,fill opacity=1 ] (104.63,236.53) -- (184.2,236.53) -- (184.2,269.87) -- (104.63,269.87) -- cycle ;

\draw  [fill={rgb, 255:red, 0; green, 91; blue, 198 }  ,fill opacity=1 ] (104.3,327.87) -- (183.87,327.87) -- (183.87,361.2) -- (104.3,361.2) -- cycle ;
\draw    (184.51,344.02) -- (204.43,344.26) ;
\draw [shift={(206.43,344.29)}, rotate = 180.7] [color={rgb, 255:red, 0; green, 0; blue, 0 }  ][line width=0.75]    (10.93,-3.29) .. controls (6.95,-1.4) and (3.31,-0.3) .. (0,0) .. controls (3.31,0.3) and (6.95,1.4) .. (10.93,3.29)   ;
\draw  [fill={rgb, 255:red, 0; green, 91; blue, 198 }  ,fill opacity=1 ] (208.63,327.87) -- (288.2,327.87) -- (288.2,361.2) -- (208.63,361.2) -- cycle ;

\draw  [fill={rgb, 255:red, 0; green, 91; blue, 198 }  ,fill opacity=1 ] (208.75,213) -- (287.87,241.38) -- (287.87,263.35) -- (208.75,291.74) -- cycle ;

\draw    (184.85,253.02) -- (204.76,253.26) ;
\draw [shift={(206.76,253.29)}, rotate = 180.7] [color={rgb, 255:red, 0; green, 0; blue, 0 }  ][line width=0.75]    (10.93,-3.29) .. controls (6.95,-1.4) and (3.31,-0.3) .. (0,0) .. controls (3.31,0.3) and (6.95,1.4) .. (10.93,3.29)   ;
\draw  [fill={rgb, 255:red, 0; green, 91; blue, 198 }  ,fill opacity=1 ] (209.08,31.67) -- (288.2,60.05) -- (288.2,82.02) -- (209.08,110.4) -- cycle ;

\draw    (185.18,71.69) -- (205.1,71.93) ;
\draw [shift={(207.1,71.95)}, rotate = 180.7] [color={rgb, 255:red, 0; green, 0; blue, 0 }  ][line width=0.75]    (10.93,-3.29) .. controls (6.95,-1.4) and (3.31,-0.3) .. (0,0) .. controls (3.31,0.3) and (6.95,1.4) .. (10.93,3.29)   ;
\draw  [fill={rgb, 255:red, 0; green, 91; blue, 198 }  ,fill opacity=1 ] (312.42,31.67) -- (391.53,60.05) -- (391.53,82.02) -- (312.42,110.4) -- cycle ;

\draw    (288.51,71.69) -- (308.43,71.93) ;
\draw [shift={(310.43,71.95)}, rotate = 180.7] [color={rgb, 255:red, 0; green, 0; blue, 0 }  ][line width=0.75]    (10.93,-3.29) .. controls (6.95,-1.4) and (3.31,-0.3) .. (0,0) .. controls (3.31,0.3) and (6.95,1.4) .. (10.93,3.29)   ;
\draw    (456.45,208.18) -- (476.37,208.42) ;
\draw [shift={(478.37,208.44)}, rotate = 180.7] [color={rgb, 255:red, 0; green, 0; blue, 0 }  ][line width=0.75]    (10.93,-3.29) .. controls (6.95,-1.4) and (3.31,-0.3) .. (0,0) .. controls (3.31,0.3) and (6.95,1.4) .. (10.93,3.29)   ;
\draw  [fill={rgb, 255:red, 0; green, 91; blue, 198 }  ,fill opacity=1 ] (479.97,191.87) -- (559.53,191.87) -- (559.53,225.2) -- (479.97,225.2) -- cycle ;

\draw (107.47,62.8) node [anchor=north west][inner sep=0.75pt]   [align=left] {\textcolor[rgb]{1,1,1}{Conv}$\displaystyle \textcolor[rgb]{1,1,1}{1\times 1}$};
\draw (107.13,153.8) node [anchor=north west][inner sep=0.75pt]   [align=left] {\textcolor[rgb]{1,1,1}{Conv}$\displaystyle \textcolor[rgb]{1,1,1}{1\times 1}$};
\draw (107.13,243.8) node [anchor=north west][inner sep=0.75pt]   [align=left] {\textcolor[rgb]{1,1,1}{Conv}$\displaystyle \textcolor[rgb]{1,1,1}{1\times 1}$};
\draw (110.8,336.13) node [anchor=north west][inner sep=0.75pt]   [align=left] {\textcolor[rgb]{1,1,1}{\textbf{AVG Pool}}};
\draw (211.13,335.13) node [anchor=north west][inner sep=0.75pt]   [align=left] {\textcolor[rgb]{1,1,1}{Conv}$\displaystyle \textcolor[rgb]{1,1,1}{1\times 1}$};
\draw (212.61,242.88) node [anchor=north west][inner sep=0.75pt]   [align=left] {\textcolor[rgb]{1,1,1}{Conv}$\displaystyle \textcolor[rgb]{1,1,1}{3\times 3}$};
\draw (212.95,61.55) node [anchor=north west][inner sep=0.75pt]   [align=left] {\textcolor[rgb]{1,1,1}{Conv}$\displaystyle \textcolor[rgb]{1,1,1}{3\times 3}$};
\draw (316.28,61.55) node [anchor=north west][inner sep=0.75pt]   [align=left] {\textcolor[rgb]{1,1,1}{Conv}$\displaystyle \textcolor[rgb]{1,1,1}{3\times 3}$};
\draw (482.47,199.13) node [anchor=north west][inner sep=0.75pt]   [align=left] {\textcolor[rgb]{1,1,1}{Conv}$\displaystyle \textcolor[rgb]{1,1,1}{1\times 1}$};

\end{tikzpicture}
    }
    \caption{The simple Inception subnetwork architecture adapted from \cite{Szegedy2017Inception-v4}. The blue-filled circle indicates concatenation operation. Filter sizes are shown in the figure. Please refer to \cite{Szegedy2017Inception-v4} for more details about the parameters of each convolution.}
    \label{fig:inception}
\end{figure}
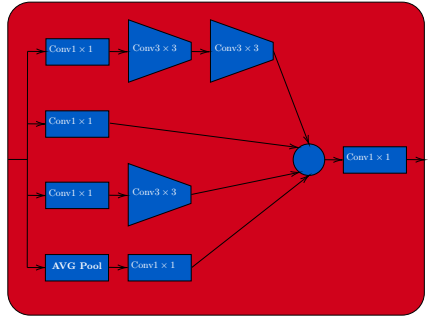

\tikzset{every picture/.style={line width=0.75pt}} 
\begin{figure}[ht]
    \centering
    \scalebox{0.5}{
        \begin{tikzpicture}[x=0.75pt,y=0.75pt,yscale=-1,xscale=1]

\draw  [fill={rgb, 255:red, 208; green, 2; blue, 27 }  ,fill opacity=1 ] (188.58,131.35) .. controls (188.58,122.42) and (195.82,115.18) .. (204.75,115.18) -- (500.65,115.18) .. controls (509.58,115.18) and (516.82,122.42) .. (516.82,131.35) -- (516.82,325.01) .. controls (516.82,333.94) and (509.58,341.18) .. (500.65,341.18) -- (204.75,341.18) .. controls (195.82,341.18) and (188.58,333.94) .. (188.58,325.01) -- cycle ;
\draw  [fill={rgb, 255:red, 0; green, 91; blue, 198 }  ,fill opacity=1 ] (235.17,232.72) -- (305.17,257.95) -- (305.17,286.22) -- (235.17,311.45) -- cycle ;
\draw  [fill={rgb, 255:red, 0; green, 91; blue, 198 }  ,fill opacity=1 ] (235.07,143.27) -- (305.07,168.5) -- (305.07,196.77) -- (235.07,222) -- cycle ;
\draw  [fill={rgb, 255:red, 0; green, 91; blue, 198 }  ,fill opacity=1 ] (354.7,228.18) .. controls (354.7,217.2) and (363.6,208.3) .. (374.58,208.3) .. controls (385.55,208.3) and (394.45,217.2) .. (394.45,228.18) .. controls (394.45,239.15) and (385.55,248.05) .. (374.58,248.05) .. controls (363.6,248.05) and (354.7,239.15) .. (354.7,228.18) -- cycle ;
\draw [fill={rgb, 255:red, 0; green, 91; blue, 198 }  ,fill opacity=1 ]   (305.13,272.05) -- (360.83,244.93) ;
\draw [shift={(362.63,244.05)}, rotate = 154.04] [color={rgb, 255:red, 0; green, 0; blue, 0 }  ][line width=0.75]    (10.93,-3.29) .. controls (6.95,-1.4) and (3.31,-0.3) .. (0,0) .. controls (3.31,0.3) and (6.95,1.4) .. (10.93,3.29)   ;
\draw [fill={rgb, 255:red, 0; green, 91; blue, 198 }  ,fill opacity=1 ]   (305.63,182.8) -- (359.12,211.85) ;
\draw [shift={(360.88,212.8)}, rotate = 208.5] [color={rgb, 255:red, 0; green, 0; blue, 0 }  ][line width=0.75]    (10.93,-3.29) .. controls (6.95,-1.4) and (3.31,-0.3) .. (0,0) .. controls (3.31,0.3) and (6.95,1.4) .. (10.93,3.29)   ;
\draw  [fill={rgb, 255:red, 0; green, 91; blue, 198 }  ,fill opacity=1 ] (424.42,188.67) -- (494.42,213.9) -- (494.42,242.17) -- (424.42,267.4) -- cycle ;

\draw [fill={rgb, 255:red, 0; green, 91; blue, 198 }  ,fill opacity=1 ]   (394.45,228.18) -- (422.1,228.27) ;
\draw [shift={(424.1,228.28)}, rotate = 180.19] [color={rgb, 255:red, 0; green, 0; blue, 0 }  ][line width=0.75]    (10.93,-3.29) .. controls (6.95,-1.4) and (3.31,-0.3) .. (0,0) .. controls (3.31,0.3) and (6.95,1.4) .. (10.93,3.29)   ;
\draw    (188.33,227.87) -- (212.58,227.87) ;
\draw    (212.58,182.87) -- (212.87,273.15) ;
\draw    (212.58,182.87) -- (232.5,183.11) ;
\draw [shift={(234.5,183.13)}, rotate = 180.7] [color={rgb, 255:red, 0; green, 0; blue, 0 }  ][line width=0.75]    (10.93,-3.29) .. controls (6.95,-1.4) and (3.31,-0.3) .. (0,0) .. controls (3.31,0.3) and (6.95,1.4) .. (10.93,3.29)   ;
\draw    (212.87,273.15) -- (232.79,273.39) ;
\draw [shift={(234.79,273.42)}, rotate = 180.7] [color={rgb, 255:red, 0; green, 0; blue, 0 }  ][line width=0.75]    (10.93,-3.29) .. controls (6.95,-1.4) and (3.31,-0.3) .. (0,0) .. controls (3.31,0.3) and (6.95,1.4) .. (10.93,3.29)   ;
\draw    (494.18,228.01) -- (514.1,228.25) ;
\draw [shift={(516.1,228.28)}, rotate = 180.7] [color={rgb, 255:red, 0; green, 0; blue, 0 }  ][line width=0.75]    (10.93,-3.29) .. controls (6.95,-1.4) and (3.31,-0.3) .. (0,0) .. controls (3.31,0.3) and (6.95,1.4) .. (10.93,3.29)   ;

\draw (426.92,218.55) node [anchor=north west][inner sep=0.75pt]   [align=left] {\textcolor[rgb]{1,1,1}{\textbf{Conv}}$\displaystyle \textcolor[rgb]{1,1,1}{1\times1}$};
\draw (237.67,250.6) node [anchor=north west][inner sep=0.75pt]   [align=left] {\begin{minipage}[lt]{43.85pt}\setlength\topsep{0pt}
\begin{center}
\textcolor[rgb]{1,1,1}{Deform}\\\textcolor[rgb]{1,1,1}{\textbf{}$\displaystyle \textcolor[rgb]{1,1,1}{3\times3}$}
\end{center}

\end{minipage}};
\draw (237.5,161.13) node [anchor=north west][inner sep=0.75pt]   [align=left] {\begin{minipage}[lt]{43.85pt}\setlength\topsep{0pt}
\begin{center}
\textcolor[rgb]{1,1,1}{Regular}\\\textcolor[rgb]{1,1,1}{\textbf{}}\textcolor[rgb]{1,1,1}{$\displaystyle \textcolor[rgb]{1,1,1}{3\times3}$}
\end{center}

\end{minipage}};

\end{tikzpicture}

    }
    \caption{The deformable inception block.}
    \label{fig:deform_inception}
\end{figure}
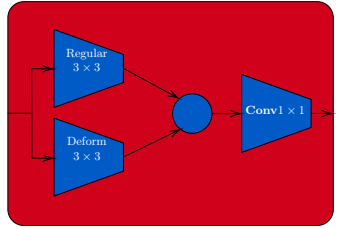

\subsection{Adopted Tracking Framework}
We hypothesize that modern ViT-based single-object trackers, such as OSTrack \cite{ye2022joint} and ODTrack \cite{zheng2024odtrack} lack good exploitation of the joint search and template embeddings to extract the location information.

Given a ViT-based backbone, the output after the joint feature extraction and relation modelling phase is an embedding tensor of dimensions $(B, H \times W, D)$ where $B$ is the batch size, $H \times W$ is the dimension of the transformer patch, and $D$ is the embedding dimension. For the ViT-based trackers represented by the three selected trackers, this output tensor is subjected to a CNN-based bounding box regression network that extracts three score maps representing the center location, the size, and the offset of the center of the bounding box as shown in Fig.~\ref{fig:overall_arch}. Given that the input tensor might contain global correlation as well as local correlation, we hypothesize it is essential that the bounding box network has multiple receptive fields that enable the network to capture relationships on both the global and local levels. 

\subsection{Inception Network}
To this end, we propose the usage of an Inception block \cite{szegedy2015going} in two alternatives as depicted in Fig.~\ref{fig:inception} and Fig.~\ref{fig:deform_inception}. 

The simpler inception block is adopted from \cite{Szegedy2017Inception-v4}. As depicted in Fig.~\ref{fig:inception}, it is composed of four parallel convolutional paths. The four convolutional paths share the functionality of mapping the input feature map to a new feature space while each of them has a different receptive field. Namely, while a bigger filter size of $7\times7$ that is realized by the two $3\times3$ convolutions shown in the upper-most branch foster a more global understanding of attentional relationships in the input tensor, the smaller ones ($1\times1$, $3\times3$) focus on a local context. After feeding the embedding tensor into the convolutional layers, the obtained outputs are concatenated and fed into a $1\times1$ convolutional filter to reduce the number of channels to the original input number of channels, hence, speeding up the computations. 

The more sophisticated inception block is based on the same idea but with a different implementation. Given the intrinsic adaptive receptive field that Deformable Convolutional Networks have \cite{Zhu2019Deformable}, we design a deformable-based inception block of convolutions that have filter size of $3\times3$ where each filter performs deformable convolution operation and the final output is concatenated and fed to a regular $1\times1$ convolution as shown in Fig.~\ref{fig:deform_inception}. The deformable convolutions enable a larger receptive field for the same filter size by learning the sampling positions of the input. In addition to the deformable convolution, we add a branch of regular convolution to consider local contiguous context, therefore simplifying the learning objective and decreasing convergence time. 

\section{Experiments}
\label{sec:experiments}

\begin{table}
\centering
\caption{Performance comparison against the state-of-the-art methods. We emphasize that our models are only trained on the training split of the GOT-10k dataset. Significant results are in \textbf{bold}}
\label{tab:sota}
\resizebox{\columnwidth}{!}{%
\begin{tabular}{c|c|c|c|c|c} 
\hline
\multirow{2}{*}{Method}               & \multicolumn{3}{c|}{GOT-10k}    & UAV123                                           & OTB2015  \\ 
\cline{2-6}
                                      & AO     & $SR_{0.5}$ & $SR_{0.75}$ & AUC                                              & AUC      \\ 
\hline
SiamFC \cite{bertinetto2016fully}     & $34.8$ & $35.3$     & $9.8$     & $46.5$                                           & $58.2$   \\
ATOM \cite{Danelljan2019ATOM}         & $55.6$ & $63.4$     & $40.2$    & $65.0$                                           & -        \\
DiMP \cite{Bhat2019Learning}          & $61.1$ & $71.7$     & $49.2$    & $65.4$                                           & $68.4$   \\
SiamRCNN \cite{Voigtlaender2020Siam}  & $64.9$ & $72.8$     & $59.7$    & $64.9$                                           & $70.1$   \\
Stark \cite{Yan2021Learning}          & $68.8$ & $78.1$     & $64.1$    & \begin{tabular}[c]{@{}c@{}}$69.1$\\\end{tabular} & $68.5$   \\
SBT-B \cite{Xie2022Correlation-Aware} & $69.9$ & $80.4$     & $63.6$    & -                                                & $70.9$   \\
Mixformer \cite{cui2022mixformer}     & $70.7$ & $80.0$     & $67.8$    & $70.4$                                           & -        \\
OSTrack~\cite{ye2022joint}            & $73.7$ & $83.2$     & $70.8$    & $\mathbf{70.7}$                                           & -        \\
AiATrack~\cite{Gao2022AiATrack}       & $69.6$ & $80.0$     & $63.2$    & $70.6$                                           & $69.6$   \\
SeqTrack \cite{Chen2023SeqTrack}      & $74.5$ & $84.3$     & $71.4$    & $68.6$                                           & -        \\
GRM \cite{Gao2023Generalized}         & $73.4$ & $82.9$     & $70.4$    & $70.2$                                           & -        \\
VideoTrack \cite{Xie2023VideoTrack}   & $72.9$ & $81.9$     & $69.8$    & $69.7$                                           & -        \\
ARTrack \cite{wei2023autoregressive}  & $75.5$ & $84.3$     & $74.3$    & $70.5$                                           & -        \\
ODTrack-B~\cite{zheng2024odtrack}     & $77.0$ & $\mathbf{87.9}$     & $75.1$    & -                                                & $\mathbf{72.3}$   \\ 
\hline
ODTrack-B (Our machine)               & $75.6$ & $86.1$     & $72.7$    & $65.8$                                           & $68.5$   \\
ODTrack-B-Inception (Ours)            & $\mathbf{77.3}$ & $87.5$     & $\mathbf{75.5}$    & $65.8$                                           & $69.0$   \\
\hline
\end{tabular}
}
\end{table}

\begin{table}[ht]
\caption{Investigation of the performance of the two proposed bounding box networks on three modern trackers. Significant results are in \textbf{bold}. All three trackers are retrained from MAE \cite{he2022masked} pretraining weights on our machine using the same configurations provided in their repositories for fair comparison.}
\label{tab:ablation}
\resizebox{\columnwidth}{!}{%
\begin{tabular}{c|ccc|c|c}
\hline
\multirow{2}{*}{Method} & \multicolumn{3}{c|}{GOT-10k}  & UAV123 & OTB2015 \\ \cline{2-6} 
                      & \multicolumn{1}{c|}{AO} & \multicolumn{1}{c|}{$SR_{0.5}$} & $SR_{0.75}$ & AUC & AUC   \\ \hline
OSTrack-B  & \multicolumn{1}{c|}{$70.1$}   & \multicolumn{1}{c|}{$79.0$}      &   $66.3$     & $65.79$  &   $66.68$  \\ 
OSTrack-B-Inception   & \multicolumn{1}{c|}{$71.6$}   & \multicolumn{1}{c|}{$80.7$}      &   $66.9$     &    $66.01$ & $68.18$   \\ 
OSTrack-B-deformable  & \multicolumn{1}{c|}{$70.8$}   & \multicolumn{1}{c|}{$79.8$}      &    $67.1$    &    $\mathbf{67.78}$  & $66.98$   \\ \hline
ODTrack-B & \multicolumn{1}{c|}{$75.6$}   & \multicolumn{1}{c|}{$86.1$}      &  $72.7$     &    $65.79$  & $68.46$   \\ 
ODTrack-B-Inception   & \multicolumn{1}{c|}{$\mathbf{77.3}$}   & \multicolumn{1}{c|}{$\mathbf{87.5}$}      &    $\mathbf{75.5}$    &   $65.79$ & $\mathbf{69.03}$   \\ 
ODTrack-B-deformable  & \multicolumn{1}{c|}{$76.8$}   & \multicolumn{1}{c|}{$87.2$}      &    $73.3$    & $64.30$  &  $68.89$  \\ \hline
\end{tabular}%
}
\end{table}

\begin{table}[ht]
\caption{Study of the performance of ODTrack with deformable convolution Inception module. "deformable only" refers to the deformable branch without the regular convolution while "Conv only" refers to the \textit{original} ODTrack \cite{zheng2024odtrack}. "Both" refers to using the bypass branch. Significant results are in \textbf{bold}.  }
\label{tab:ablation-deformable}
\resizebox{\columnwidth}{!}{%
\begin{tabular}{c|ccc}
\hline
\multirow{2}{*}{Method} & \multicolumn{3}{c}{GOT-10k}  \\ \cline{2-4} 
                        & \multicolumn{1}{c|}{AO($\%$)} & \multicolumn{1}{c|}{$SR_{0.5}(\%)$} & $SR_{0.75}(\%)$    \\ \hline
deformable only         & \multicolumn{1}{c|}{$75.4$}   & \multicolumn{1}{c|}{$86.6$}      &  $\mathbf{73.3}$         \\ 
Conv only               & \multicolumn{1}{c|}{$75.6$}   & \multicolumn{1}{c|}{$86.1$}      &  $72.7$          \\ 
Both                    & \multicolumn{1}{c|}{$\mathbf{76.8}$}   & \multicolumn{1}{c|}{$\mathbf{87.2}$}      &    $\mathbf{73.3}$     \\ \hline
\end{tabular}%
}
\end{table}

\subsection{Implementation Details}
We adopt the ViT-base \cite{dosovitskiy2021an} as the visual encoder with MAE pretraining weights \cite{he2022masked} to train OSTrack \cite{ye2022joint} and ODTrack \cite{zheng2024odtrack} using our proposed bounding box regression networks. The template image shape is $192\times192\times3$ while the search image shape is $384\times384\times3$. All models are trained for $100$ epochs solely on the GOT-10k dataset \cite{Huang2021GOT-10K}. Training is conducted on a server with two $50$ GB GPUs. The batch size is set to $8$. AdamW \cite{loshchilov2018decoupled} is used to optimize the training objective. We sample $40000$ examples per epoch to train networks based on ODTrack following the configurations found in their code repository, while the number of samples per epoch for OSTrack-based networks is $60000$. We keep the learning rate and the weight decay that each model was originally trained with. We report the performance of the two proposed variants, namely the Inception module and the Deformable Inception module. The basic convolutional block used in both of them is a stack of conv, ReLU, and BatchNorm. The filter details of the convolutions for each variant are shown in Fig.~\ref{fig:inception} and Fig.~\ref{fig:deform_inception}.

\subsection{State of the Art Comparison}
We compare our method to $14$ methods on three leading datasets, GOT-10k \cite{Huang2021GOT-10K}, UAV123 \cite{Mueller2016A} and OTB2015 \cite{Wu2015Object}. The results shown in Table~\ref{tab:sota} indicate that our method variants outperform the state-of-the-art techniques on the three datasets when compared with retrained models on the same machines.

\subsubsection{GOT-10k} Generic Object Tracking dataset is a large-scale single object tracking dataset. It contains $9334$ training sequences, $180$ validation sequences and $180$ test sequences. The testing protocol of the dataset requires models to be trained only on the GOT-10k training sequences. The training and test sequences are not of overlapping classes. As reported in Table~\ref{tab:sota}, our best performant variant outperforms the state-of-the-art ODTrack-base \cite{zheng2024odtrack} trained on our machine by $1.7\%$ in Average Overlap (AO). Moreover, for the Success Rate at $0.5$ IoU threshold ($SR_{0.5}$) and the ($SR_{0.75}$), our model outperforms ODTrack by $1.4\%$ and $2.8\%$ respectively. 

\subsubsection{UAV123} UAV123 is a aerial viewpoint video dataset that contains $123$ challenging sequences. It contains training and testing sequences. However, we only use the testing set for evaluation purposes without training on the training set. According to the shown results in Table~\ref{tab:sota}, our method outperforms the aligned state-of-the-art techniques while it is only trained on the GOT-10k training split. 

\subsubsection{OTB2015} Object Tracking Benchmark \cite{Wu2015Object} is a pioneering testing set for single object tracking that consists of $100$ challenging frames collected from YouTube. The dataset labels are publicly available. The performance pattern on this dataset is similar to the one in the GOT-10k as they are both datasets of the same natural image domain.

\subsection{Ablation Study}
We examine the effect of the proposed simple Inception and deformable convolution Inception on the two adopted modern trackers stated in section \ref{sec:methodology}. As reported in Table~\ref{tab:ablation}, our Inception and deformable inception methods outperform the locally trained models on the GOT-10k test set considering the Average overlap by $1.5\%$ and $0.6\%$ respectively. The pattern is consistent despite the different tracking ideas implemented in the different modules, indicating the effectiveness of the inception and deformable inception convolutional networks in learning a suitable receptive field to process the joint search and template embeddings. Notably, the pure Inception is more powerful enhancing the performance on all three datasets. This indicates the higher learning capacity of the inception module compared to the deformable inception. We also investigate the effect of the skip branch that has the $3\times3$ convolutional filter of the deformable convolution Inception module in Table
~\ref{tab:ablation-deformable}. It can be shown that the adopted variant is the best performant deformable module indicating the importance of the $3\times3$ convolution skip connection in facilitating the learning objective and increasing model capacity.

\subsection{Visualization}
\begin{figure}
    \centering
    \scalebox{0.2}{
        \includegraphics[]{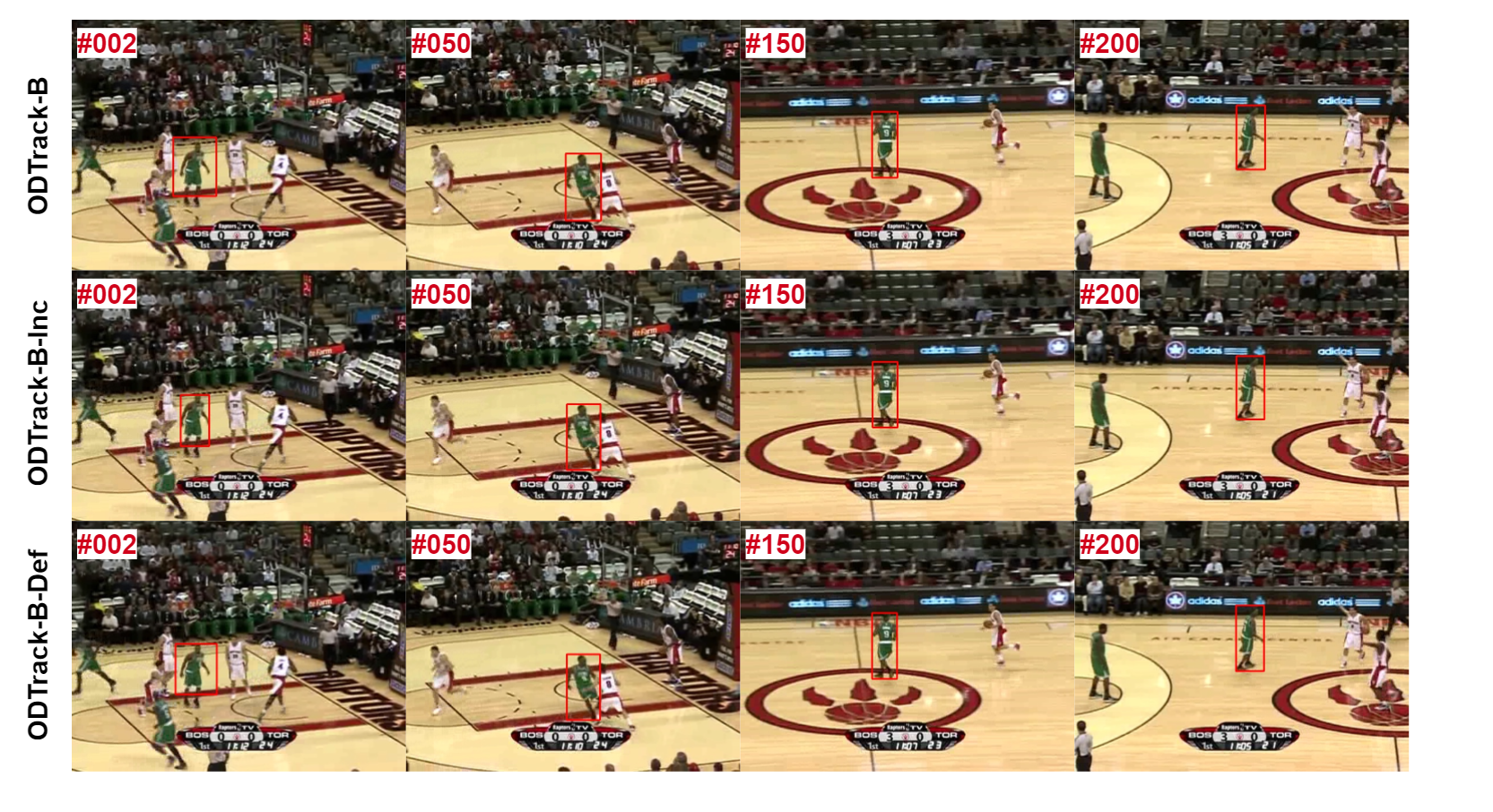}
    }
    \caption{Success cases of the proposed methods on the OTB2015 Basketball sequence. It can be noted that the pure Inception network is most accurate especially when comparing the first column of frames. Although bounding boxes converge at frame $\#50$, the initially predicted bounding boxes at frame $\#2$ indicate the fast adaptation of the Inception bounding box network.}
    \label{fig:vis-success}
\end{figure}

We qualitatively visualize the output of our two bounding box regression networks in Fig.~\ref{fig:vis-success}. As depicted, the regular Inception has a better bounding box quality. This indicates the high model capacity of the inception module that enables the best exploitation of the visual features given by the ViT in ODTrack \cite{zheng2024odtrack}.

\section{Conclusion}
\label{sec:conclusion}
In this work, we show the significance of the bounding box regression given the visual interaction features of a one-stream tracking transformer. We show that a bounding box-regression network lacking a suitable receptive field can hinder the performance of a transformer tracking model due to the lack of good exploitation of the visual features provided by the backbone. Experiments show that our methods of Inception and deformable inception bounding box networks perform favourably against state-of-the-art ODTrack by $1.7\%$ of AO on the GOT-10k \cite{Huang2021GOT-10K} dataset. As future pathways, this work can be extended to ablate the different emerging CNN techniques and find the CNN bounding box network that strikes the balance between a good inductive bias and a global receptive field.

\begin{credits}

\subsubsection{\discintname}The authors have no competing interests to declare that are
relevant to the content of this article.
\end{credits}
%
%
%
\bibliographystyle{splncs04}

%





\end{document}